\documentclass[a4paper,fleqn]{cas-dc}

\usepackage[authoryear,longnamesfirst]{natbib}

\usepackage{amssymb}
\usepackage{amsmath}

\usepackage{graphicx}
\usepackage[misc]{ifsym}
\usepackage{subcaption}

\usepackage{microtype}

\usepackage{natbib}
\setcitestyle{numbers,square}

\def\tsc#1{\csdef{#1}{\textsc{\lowercase{#1}}\xspace}}
\tsc{WGM}
\tsc{QE}

\begin{document}
\let\printorcid\relax
\let\WriteBookmarks\relax
\def\floatpagepagefraction{1}
\def\textpagefraction{.001}

\shorttitle{}    

\shortauthors{Xinru Meng et al}  

\title [mode = title]{Energy-Based Pseudo-Label Refining for Source-free Domain Adaptation}  



\author[label1]{Xinru Meng} 
\ead{mxrmemm@nuaa.edu.cn}
\credit{Conceptualization, Methodology, Investigation, Formal analysis, Writing - original draft}

\author[label1]{Han Sun} 
\cormark[1]
\ead{sunhan@nuaa.edu.cn}
\credit{Conceptualization, Funding acquisition, Methodology, Supervision, Writing - review \& editing}

\author[label1]{Jiamei Liu} 
\ead{jiamei@nuaa.edu.cn}
\credit{Data curation, Writing - original draft}

\author[label1]{Ningzhong Liu} 
\ead{liunz@163.com}
\credit{Visualization, Investigation}

\author[label2]{Huiyu Zhou} 
\ead{hz143@leicester.ac.uk}
\credit{Resources, Supervision}

\affiliation[label1]{organization={School of Computer Science and Technology},
            addressline={Nanjing University of Aeronautics and Astronautics}, 
            city={Nanjing},
            postcode={211100}, 
            state={Jiangsu},
            country={China}}
\affiliation[label2]{organization={School of Computing and Mathematical Sciences},
            addressline={University of Leicester}, 
            city={Leicester},
            postcode={LE1 7RH}, 
            state={Capital of Leicestershire},
            country={UK}}















\cortext[1]{Corresponding author}



\begin{abstract}
Source-free domain adaptation (SFDA), which involves adapting models without access to source data, is both demanding and challenging. Existing SFDA techniques typically rely on pseudo-labels generated from confidence levels, leading to negative transfer due to significant noise. To tackle this problem, Energy-Based Pseudo-Label Refining (EBPR) is proposed for SFDA. Pseudo-labels are created for all sample clusters according to their energy scores. Global and class energy thresholds are computed to selectively filter pseudo-labels. Furthermore, a contrastive learning strategy is introduced to filter difficult samples, aligning them with their augmented versions to learn more discriminative features. Our method is validated on the Office-31, Office-Home, and VisDA-C datasets, consistently finding that our model outperformed state-of-the-art methods.
\end{abstract}





\maketitle

\section{Introduction}
Supervised learning models trained on the original domain often struggle to generalize effectively to new domains due to differences in data distribution. Domain adaptation (DA)~\cite{wilson2020survey, chaddad2023domain} addresses this challenge by aligning the data distributions of the source and target domains. However, accessing source data may raise privacy and intellectual property issues during adaptation, which complicates the knowledge transfer process. Therefore, Source-Free Domain Adaptation (SFDA) has received much attention for its practicality.

To address the challenges of SFDA, the most widely used technique in self-supervised learning is pseudo-label. However, existing pseudo-label methods typically consist of two steps, each with its limitations. 1) Pseudo-label generation: The model prediction outputs of the samples are directly used as pseudo-labels. This approach is prone to assigning incorrect categories due to distributional shifts. 2) Pseudo-label filtering: This involves using the output probabilities of the categories to determine whether the pseudo-labels should be filtered. However, the output of the source domain model still retains the results of the source domain distribution. Therefore, the selected reliable pseudo-labels are of limited help to the adaptation process. In addition, the filtering process is often affected by the "winner-take-all" problem, where the model overfits the majority class and ignores the minority class.

To effectively address the above challenges, we propose a new framework aimed to generate more accurate pseudo-labels. The main difference between our proposed framework and benchmark methods such as SHOT\cite{liang2020we}, A2Net\cite{xia2021adaptive}, and DIPE\cite{wang2022exploring} is that EBPR generates pseudo-labels through energy scores and adaptively filters pseudo-labels using global and local energy thresholds, while also introducing a contrast learning strategy to handle challenging samples. In contrast, SHOT mainly relies on soft labels and entropy minimization, A2Net focuses on adversarial training and feature alignment, and DIPE focuses on distribution matching and class balancing. Without relying on the source data, EBPR pays more attention to generating reliable pseudo-labels from the energy level, and gets more refined pseudo-label through the energy threshold to reduce the impact of noise on model adaptability.

We integrate three well-designed modules into EBPR: pseudo-label generation with transportation cost (PGTC), adaptive energy threshold (AET), and enhanced consistency training (ECT). First, inspired by the consistency of different data distributions in the two domains at the energy measurement level, PGTC introduces the energy score as a clustering element for all samples. Second, since the majority class heavily influences the final class of a cluster, AET designs an adaptive energy threshold to address the pseudo-label category imbalance problem. The intuition behind AET is that both global and category energy scores should be considered to adaptively filter out false pseudo-labels in different classes. In addition, for harder samples without reliable pseudo-labels in AET, we introduce consistency learning between different views to further mine target data features. Our main contributions are summarized below:
\begin{itemize}
 \item We tackle the challenging SFDA task with a novel framework, namely energy-based pseudo-label refining method (EBPR). The method reduces the domain gap by using pseudo-label refining.
 \item We highlight some of the important aspects of EBPR as: a) generating reliable pseudo-labels at the energy level, b) designing an adaptive filtering threshold for pseudo-labels through the energy score, and c) introducing consistent learning for hard samples to learn specific features.
 \item We establish the efficacy of EBPR through extensive experiments on three benchmark datasets.

\end{itemize}

\section{Related Works}
\subsection{Pseudo-labels methods}
The pseudo-label method aims to obtain high-quality pseudo-labels through a process involving pseudo-label generation and filtering. Primarily, the generation step assigns pseudo-labels based on the similarity between the target sample and the class prototype. BMD~\cite{qu2022bmd} introduces a dynamic pseudo-label strategy for updating pseudo-labels during the domain adaptation process. Qiu et al.~\cite{qiu2021source} generates feature prototypes for each source category using a conditional generator to create pseudo-label for the target domain data. However, these methods generate pseudo-labels based on confidence levels or feature clustering, introducing considerable noise due to domain shifts. The latter typically designs a specific rule mechanism to preserve label purity. For instance, Kim et al.~\cite{kim2021domain} proposes when it meets certain conditions regarding its distances from the most similar and the second similar class prototype. In contrast to these methods, EBPR utilizes the energy score as a clustering guide to produce pseudo-labels that ensure consistency between the two domains. ~\cite{fofanah2025chamformer} and ~\cite{fofanah2024stalformer} propose multivariate feature-aware learning and augmented feature learning to explore the model's prediction of samples in more dimensions.
\subsection{Energy-based models}
 The core concept of an energy-based model (EBM)  is to establish a dependency that assigns a corresponding scalar (energy) to every pair of variables in the input space. Observing that the disparity in data distribution can be accurately characterized by free energy~\cite{lecun2006tutorial}. Samples with inconsistent class distributions between the two domains have a higher degree of matching on the energy scale. This inspired EBPR to optimize the quality of pseudo-label. ~\cite{xie2022active} have introduced an energy-based approach. This method proficiently identifies distinctive samples from the target domain and diminishes the difference between source and target domains by directly minimizing the energy difference between them.

\section{Proposed Methodology}
\subsection{Problem definition}
In a source-free domain adaptation task, the labeled source domain data with $n_s$ labeled samples is denoted as $D_s = \{{(x_s^{i},y_s^{i})}\}_{i=1}^{n_s}$, where $x_s^{i} \in \mathcal{X}_s$,${y}_s^{i} \in \mathcal{Y}_s \subseteq \mathbb{R}^K$ and $y_s^{i}$ is the one-hot ground-truth label with $K$ dimensions. 
 The unlabeled target domain data with $n_t$ samples is denoted as $D_t = \{{x_t^{i}}\}_{i=1}^{n_t}$, where $x_t^{i} \in \mathcal{X}_t$. The label space of the two domains is shared while the marginal distribution is different. The source model, denoted as $f_s(\cdot)=h_s(g_s(\cdot))$, consists of feature extractor $g_s(\cdot):\mathcal{X}_s\to \mathbb{R}^D$ and classifier $h_s(\cdot):\mathbb{R}^D\to \mathbb{R}^K$ where $D$ is the feature dimension. SFDA aims to predict the labels $\{y_t^{i}\}_{i=1}^{n_t}$ in the target domain where $y_t^{i} \in  \mathcal{Y} \subseteq \mathbb{R}^K$  by learning a target model $f_t(\cdot):\mathcal{X}_t \to \mathcal{Y}_t$ based on the source model. Similar to the formula for the source model, the target model is defined as $f_t(\cdot)=h_t(g_t(\cdot))$.

\subsection{Method}
\subsubsection{Overview of the training process} 
\begin{figure}[!t]
\centering
\includegraphics[width=0.5\textwidth]{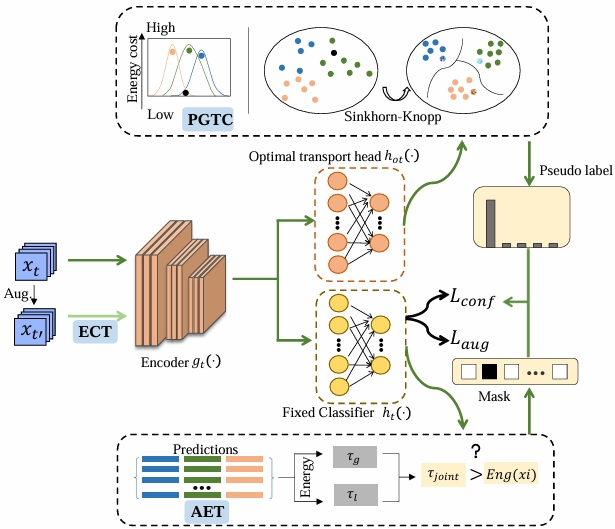}
\caption{\footnotesize The framework of the proposed EBPR.} \label{framework1}
\vspace{-0.2em}
\end{figure}

 The architecture of the EBPR network, shown in Fig. \ref{framework1}, is structured around three core modules. Initially, the feature encoder module, denoted as $g_t(\cdot)$, processes the target domain data to extract the feature vectors. These vectors are then fed into the optimal transport head $h_{ot}$, which computes the mapping outputs. By using the mapping outputs to calculate the energy score for each sample, we can obtain the clustering distribution for all samples based on the energy score. The pseudo-label of each sample can be obtained on the basis of the clustering distribution. At the same time, feature vectors are also passed through a classifier to generate classification outputs, which are used to generate the energy score. Then we introduce an adaptive threshold based on energy score used for pseudo-label filtering and divide the target samples into samples of low-reliability and samples with high-reliability. For high-reliability samples, the classification loss is constructed between pseudo-labels obtained by PGTC and the model's predicted outputs to optimize the target model. Samples with low-reliability are managed by imposing constraints that enhance the consistency of the local structure within the embedded space, ensuring a more robust representation of the data.
The overall loss of EBPR is:
\begin{equation}
L = L_{\rm conf} + \rho L_{\rm aug},
  \label{eq3}
\end{equation}
where $\rho$ is the trade-off parameters.

\subsubsection{Pseudo-label generation with transport cost} 
Domain adaptation can be articulated as a mathematical challenge where one distribution is transformed into another.  Optimal Transport (OT)  ~\cite{lu2023uncertainty} aims to transform one probability distribution into another with the minimal cost. In closed-set unsupervised classification tasks, OT can assign pseudo-labels to unlabeled samples. In the above process, the goal is to minimize the transformation cost, which requires obtaining an optimal transport matrix. The Sinkhorn-Knopp algorithm is an iterative method specifically designed for computing the aforementioned optimal transport matrix. This algorithm, through its iterative process, effectively finds an optimal transport plan that satisfies specific cost constraints.

Therefore, the PGTC module introduces a linear layer of $M$ dimension as the optimal transport header $h_{ot}$, mapping the embedding vector to $C$ clusters. Then we can define the cost matrix in optimal transport problem as $\hat P \in \mathbb{R}^{M \times N}$ and $P_{ji}=P(c_j|x_i)$ denotes the probability of $x_i$ belonging to $c_j$, where $c_j$ is the j-th cluster. The posterior probability of assigning samples to each cluster is defined as $Q_{ji}=q(c_j|x_i)$, and the assignment matrix is $Q \in \mathbb{R}^{M \times N}$ which represents the assignment of a certain cluster to each sample. The optimal transport scheme for assigning samples to c clusters is defined as:
\begin{equation}
U(a,b):= \{\textbf{Q} \in \mathbb{R}^{M \times N} | \textbf{Q1}_N=a, \textbf{Q1}_M=b, \textbf{Q} >0 \},
  \label{eq4}
\end{equation}
where $a$ and $b$ are the marginal distribution of the rows and columns of the matrix $\textbf{Q}$, respectively. $\textbf{1}$ is an all-ones vector.

Sample with a low energy score is more reliable, and vice versa. The energy score $\textbf{Eng}$ of each sample $x_i$ distributed across all clusters and the energy cost matrix ${\textbf{P}} \in \mathbb{R}^{M \times N}$ in the optimal transport problem are calculated, $\textbf{Eng}$ is a general reference to $Eng_i$:
\begin{equation}
Eng_i= {\rm log}\sum_{j=1}^Me^{l(c_j|x_i)},
\textbf{P}=\hat{\textbf{P}} \cdot \textbf{Eng} \in R^{M \times N}
  \label{eq5}
\end{equation}
where $l(c_j|x_i)=h_{ot}(z_i)$ represents the logit of the sample $x$ belonging to the cluster $c_j$. The elements in ${\textbf{P}}$ represent the probability of a sample being assigned to a certain cluster. Larger element values indicate that the cost of transferring samples to the corresponding cluster is lower.

A low energy score for a sample indicates that the sample has a large uncertainty in the clustering distribution, so we distribute it uniformly over the C clusters. The sample with a higher energy score is encouraged to be assigned to the corresponding cluster. 
We introduce the entropic regularization term $H(Q)=\sum_{ji}Q_{ji}logQ_{ji}$ into Wasserstein distance. The optimal problem solution is:
\begin{equation}
\begin{aligned}
{\rm OT}(a,b)
&=\underset{{Q \in U(a,b)}}{{ \textbf{min}}}- \langle \textbf{Q},\textbf{P} \rangle + \varepsilon \sum_{ji}Q_{ji}logQ_{ji},
  \label{eq6}
  \end{aligned}
\end{equation}
where $\varepsilon >$ 0, $\langle\cdot,\cdot\rangle$ represents the Fibonacci point multiplication, and $Q_{ji}$ is denoted by the posterior probability of the sample $x_i$ being assigned to the cluster $c_j$. By transforming this equation, the $\textbf{Q}$ can be denoted as $Q^{\frac{1}{\varepsilon}}={\rm Diag}(\textbf{u})\textbf{P}{\rm Diag}(\textbf{v})$. $\textbf{u}$ and $\textbf{v}$ can be solved iteratively and updated by using the Sinkhorn algorithm~\cite{distances2013lightspeed}:
\begin{equation}
\textbf{u}^{t+1}=\frac{a}{\sum_iP_{ij}\textbf{v}_i^t}, \quad
\textbf{v}^{t+1}=\frac{b}{\sum_jP_{ij}\textbf{u}_j^{t+1}},
  \label{eq7}
\end{equation}
where $P_{ij}$ denotes the probability that the sample $x_i$ belongs to the cluster $c_j$. Finally, by the above, we can get the optimal assignment matrix $\textbf{Q}$ which maps all samples onto C clusters.


After each sample is assigned to a cluster, we use majority voting to make the distribution of pseudo-labels more equitable. Specifically, we select the category with the most predicted results in the cluster as the pseudo-label $\{\hat{y}_i^t\}_{i=1}^{n_t}$ for all samples in the current cluster.

\subsubsection{Adaptive energy threshold}
SFDA has no explicit supervision signal during adaptation. This means that the problem of error accumulation can easily occur: if a sample is initially misidentified and continues to learn in self-training, it also be difficult to classify it correctly thereafter. Therefore further methods need to be used to select reliable pseudo-labels. If a fixed threshold is used to select pseudo-labels, a threshold that is too low is ineffective, and too high ignores useful information. Due to the presence of domain shift between the source and target domain data, it is not appropriate to use the model's predicted probability values directly for filtering. The energy score reflects the probability density of the sample~\cite{liu2020energy}. A lower energy score for a sample indicates that the sample is closer to the predictive distribution of the model. It is calculated from the model's output probability values but is a non-probability scalar. We design an adaptive threshold based on the energy score, called adaptive energy threshold. It calculates jointly from the global energy threshold and class energy threshold.

The initial global energy threshold $\tau_g^0$ is the global energy mean of all samples. In subsequent iterations, the global energy threshold is calculated by the formula:
\begin{equation}
\tau_g^\nu =
\left \{ 
\begin{aligned}
     &\tau_g^0, & {\rm if} \quad \nu=0,  \\
     &\lambda\tau_g^{\nu-1} + (1-\lambda) \frac{1}{N}\sum_{i=1}^N E_i,  &  {\rm otherwise,}
\end{aligned}
\right.
  \label{eq8}
\end{equation}
where $\nu$ denote the $\nu$-th iteration, $E_i = {\rm log}\sum_{j=1}^Ke^{f_t(x_t)}$ denotes the energy score of the sample $x_t$, $\lambda \in (0,1)$ is the momentum decay of the EMA.

\begin{table*}[!ht]
\centering
\caption{\footnotesize Classification accuracy (\%) on the \textbf{Office-31} dataset. The best and second-best results are shown in bold and with underline, respectively. }\label{office-31}
\begin{tabular}{|l|l|l|l|l|l|l|l|l|}
\hline
Method & A→D & A$\rightarrow$\textup  W & D$\rightarrow$\textup    A & D$\rightarrow$\textup   W & W$\rightarrow$\textup  A & W$\rightarrow$\textup D & Avg.\\
\hline
Source-only  & 80.9 & 75.6 & 59.5 & 92.8 & 60.5 & 99.2  & 78.1\\
\hline
DANN\cite{ganin2016domain}         & 79.7 & 82.0 & 68.2 & 96.9 & 67.4 & 99.1  & 82.2 \\
    CDAN\cite{long2018conditional}       & 92.9 & 94.1 & 71.0 & 98.6 & 69.3 & \textbf{100.0} & 87.7 \\
    SRDC\cite{tang2020unsupervised}        & 95.6 & \underline{95.7} & \textbf{76.7} & \textbf{99.2} & 77.1 & \textbf{100.0} & \underline{90.8} \\
    BDG\cite{yang2020bi}        & 93.6 & 93.6 & 73.2 & \underline{99.0} & 72.0 & \textbf{100.0} & 88.5 \\
    MCC\cite{jin2020minimum}      & 94.4 & 95.5 & 72.9 & 98.6 & 74.9 & \textbf{100.0} & 89.4 \\
\hline

SHOT\cite{liang2020we}   & 94.0 & 90.1 & 74.7 & 98.4 & 74.3 & \underline{99.9}  & 88.6   \\
SFDA\cite{kim2021domain}   & 92.2 & 91.1 & 71.0 & 98.2 & 71.2 & 99.5  & 87.2   \\
    BAIT\cite{yang2020unsupervised}    & 91.0 & 93.0 & 75.0 & \underline{99.0} & 75.3 & \textbf{100.0} & 88.9   \\
    A2Net\cite{xia2021adaptive}    & 94.5 & 94.0 & \textbf{76.7} & \textbf{99.2} & 76.1 & \textbf{100.0} & 90.1     \\
    HCL\cite{huang2021model}  & 94.7 & 92.5 & 75.9 &98.2 &\textbf{77.7} &\textbf{100.0}  & 89.8 \\
    AAA\cite{li2021divergence}   & 95.6 & 94.2 & 75.6 & 98.1 & 76.0 & 99.8 & 89.9   \\
    DIPE\cite{wang2022exploring}    & \underline{96.6} & 93.1 & 75.5 & 98.4 & \underline{77.2} & 99.6  & 90.1   \\
    PLUE\cite{wang2022source}    & 96.4 & 92.5 & 74.5 & 98.3 & 72.2 & \textbf{100.0}  & 89.0   \\
    ELR\cite{yi2022source}    & 93.8 & 93.3 & 76.2 & 98.0 & 76.9 & \textbf{100.0}  & 89.6   \\
\hline
\textbf{EBPR}& \textbf{97.0} & \textbf{96.9} & \underline{76.4} & \textbf{99.2} & 77.0 & \textbf{100.0} & \textbf{91.1}\\
\hline
\end{tabular}
\end{table*}

There are large differences in energy between classes, accounting for diversity within classes and imbalance between classes. AET calculates a class-specific class energy threshold. $\tau_l^0$ is the mean of all samples based on class at the first iteration. In subsequent iterations, it is also updated by EMA:
\begin{equation}
\tau_l^\nu =
\left \{ 
\begin{aligned}
     &\tau_l^0, & {\rm if} \quad \nu=0, \\
     &\lambda\tau_l^{\nu-1} + (1-\lambda) log(\sum_{i=1}^Ne^{f_t(x_i)}), & {\rm otherwise,}
\end{aligned}
\right.
  \label{eq9}
\end{equation}
where $\tau_l^\nu=[\tau_l^\nu (1),\tau_l^\nu (2),\dots,\tau_l^\nu (K)]$ is the list which contains energy scores for each class.

By integrating the global energy threshold and the class energy threshold, the final adaptative energy threshold can be calculated as follows in Eq.\ref{eq10}:
\begin{equation}
\begin{aligned}
\tau_{\rm joint}(k) &= -{\rm MaxNorm}(\tau_l(k)) \cdot \tau_g \\
&=-\frac{\tau_l(k)}{{\rm max}\{\tau_l(l):k\in[K]\}} \cdot \tau_g,
  \label{eq10}
\end{aligned}
\end{equation}
where ${\rm MaxNorm(\cdot)}$ is the maximum normalization. The target domain samples can be divided into reliable and hard samples:
\begin{equation}
\begin{aligned}
&\mathcal{X}_{conf} = \{x_i | E(x_i) < \tau_{joint}\}, \\
&\mathcal{X}_{hard} = \{x_j | E(x_j) > \tau_{joint}\}.
\end{aligned}
  \label{eq11}
\end{equation}

For reliable samples and pseudo-labels $\hat{y}_i$ generated by PGTC, the model can be further optimized and updated using cross-entropy loss as follows:
\begin{equation}
L_{conf} = -{\mathbb E}_{x \in X_{conf}}\sum_{k=1}^K \textbf{1}_{[\hat{y}_i=k]}log\delta_k(f_t(x)),
  \label{eq12}
\end{equation}
where $\textbf{1}(\cdot)$ is the indication function, and $\delta_k(\cdot)$ denotes the k-th class model prediction probability value of the sample.

\subsubsection{Enhanced consistency training} 
For challenging samples where pseudo-labels are not reliable, it may not be feasible to directly utilize pseudo-labeling for supervised training. The sample feature representations should remain invariant to small changes.
We make two different augmentations of the sample $x$ and obtain two augmentation views $x_i=t(x_i)$ and $x_i'=t'(x_i)$, whose corresponding feature representations are denoted as  $z_i = g_t(x_i)$ and $z_i'=g_t(x_i')$. The contrastive loss is constructed for each batch samples without reliable pseudo-label. Specifically, the different augmentation views of the same image are treated as positive pairs and the other $2(N-1)$ augmentation examples of each batch are treated as negative pairs. The loss function for augmentation consistency is defined as:
\begin{equation}
L_{\rm aug}=-{\rm log}\frac{{\rm exp}({\rm sim}(z_i,z_i')/ \tau)}{\sum_{k=1}^{2N}\textbf{1}_{[k \neq i]}{\rm exp}({\rm sim}(z_i,z_i')/ \tau)},
  \label{eq14}
\end{equation}
where ${\rm sim}(\alpha,\beta)$ is the cosine similarity and $\tau$ denotes the temperature parameter.

\section{Experimental Evaluations}

\subsection{Datasets}
Our method is evaluated on three benchmark datasets. \textbf{Office-31} \cite{saenko2010adapting} is composed of 4,652 images from 31 categories of everyday object consisting of 3 subsets: Amazon (\textbf{A}), Webcam (\textbf{W}), and DSLR (\textbf{D}). \textbf{Office-Home} \cite{venkateswara2017deep} is a challenging medium-sized benchmark with 65 categories and a total of 15,500 images. The dataset consists of 4 domains: Art (\textbf{A}), Clipart (\textbf{C}), Product (\textbf{P}), and Real-world (\textbf{R}). \textbf{VisDA-C} \cite{peng2017visda} is another challenging and large-scale dataset with 12 categories in two domains. Its source domain contains 152k synthetic images while the target domain has 55k real object images.

\subsection{Model architecture and experimental protocols}
All experiments are conducted using PyTorch. Following previous methods, the ResNet is utilized as the feature encoder. The optimal transport header consists of a 1-layer MLP with 128 dimensions. During the adaptation stage, the Adam optimizer is applied with a learning rate of 1e-3 for the backbone and 1e-2 for projection. Training epochs are set to 100, 40, and 2 for Office, Office-Home, and VisDA, respectively. For hyper-parameters, we set $\rho$ in Eq. (\ref{eq3}) as 0.8. All the results are averaged over three runs with seeds $\in$ {0, 1, 2}.
\subsection{Comparison to the literature and baselines}
We compare our method with three different experimental settings in Table 1-3: i) Source-only: ResNet \cite{he2016deep}; ii) Method of Domain Adaptation: unsupervised domain adaptation with source data: DANN\cite{ganin2016domain}, CDAN\cite{long2018conditional}, SRDC\cite{tang2020unsupervised}, MCC\cite{jin2020minimum}, BDG\cite{yang2020bi}, BNM \cite{cui2020towards}, SWD\cite{lee2019sliced} and PAL\cite{hu2020panda}; iii) source-free unsupervised domain adaptation: SHOT\cite{liang2020we}, SFDA\cite{kim2021domain}, BAIT\cite{yang2020unsupervised}, CPGA\cite{qiu2021source},  A2Net\cite{xia2021adaptive}, HCL\cite{huang2021model}, AAA\cite{li2021divergence}, and DIPE\cite{wang2022exploring}, PLUE\cite{wang2022source}.

\begin{table*}[!t]
\centering
\caption{\footnotesize Classification accuracy (\%) on the \textbf{Office-Home} dataset.}\label{Office-Home}
\resizebox{\textwidth}{!}{
\begin{tabular}{|l|l|l|l|l|l|l|l|l|l|l|l|l|l|}
\hline
Methods & \multicolumn{3}{c|}{A$\rightarrow$}  &  \multicolumn{3}{c|}{C$\rightarrow$} &\multicolumn{3}{c|}{P$\rightarrow$} &\multicolumn{3}{c|}{R$\rightarrow$} & Avg.\\
\cline{2-13}
~ &  C & P & R & A & P & R & A & C & R & A & C & P & \\
\hline
Source-only   & 38.1 & 58.9 & 69.4 & 48.3 & 57.7 & 61.1 & 49.4 & 36.4 & 69.9 & 64.7 & 43.7 & 75.4 & 56.1          \\
\hline
DANN\cite{ganin2016domain}         & 45.6 & 59.3 & 70.1 & 47.0 & 58.5 & 60.9 & 46.1 & 43.7 & 68.5 & 63.2 & 51.8 & 76.8 & 57.6          \\
    CDAN\cite{long2018conditional}           & 50.7 & 70.6 & 76.0 & 57.6 & 70.0 & 70.0 & 57.4 & 50.9 & 77.3 & 70.9 & 56.7 & 81.6 & 65.8          \\
    SRDC\cite{tang2020unsupervised}          & 52.3 & 76.3 & 81.0 & \underline{69.5} & 76.2 & 78.0 & \underline{68.7} & 53.8 & 81.7 & \textbf{76.3} & 57.1 & \underline{85.0} & 71.3          \\
    BNM\cite{cui2020towards}           & 52.3 & 73.9 & 80.0 & 63.3 & 72.9 & 74.9 & 61.7 & 49.5 & 79.7 & 70.5 & 53.6 & 82.2 & 67.9          \\
    
\hline
SHOT\cite{liang2020we}          & 57.1 & 78.1 & 81.5 & 68.0 & 78.2 & 78.1 & 67.4 & 54.9 & 82.2 & 73.3 & 58.8 & 84.3 & 71.8          \\
SFDA\cite{kim2021domain}          & 48.4 & 73.4 & 76.9 & 64.3 & 69.8 & 71.7 & 62.7 & 45.3 & 76.6 & 69.8 & 50.5 & 79.0 & 65.7          \\
    BAIT\cite{yang2020unsupervised}         & 57.4 & 77.5 & \underline{82.4} & 68.0 & 77.2 & 75.1 & 67.1 & 55.5 & 81.9 & 73.9 & 59.5 & 84.2 & 71.6          \\
    A2Net\cite{xia2021adaptive}        &\underline{58.4} & 79.0 & \underline{82.4} & 67.5 & \underline{79.3}& 78.9 & 68.0 & 56.2 & 82.9 & 74.1 & 60.5 & \underline{85.0} & \underline{72.8}     \\
    CPGA\cite{qiu2021source}          & \textbf{59.3} & 78.1 & 79.8 & 65.4 & 75.5 & 76.4 & 65.7 & \textbf{58.0} & 81.0 & 72.0 & \textbf{64.4} & 83.3 & 71.6          \\
    AAA\cite{li2021divergence}          & 56.7 & 78.3 & 82.1 & 66.4 & 78.5 & \underline{79.4} & 67.6 & 53.5 & 81.6 &74.5 & 58.4 & 84.1 & 71.8 \\
    DIPE\cite{wang2022exploring}         & 56.5 & \underline{79.2} & 80.7 & \textbf{70.1} & \textbf{79.8} & 78.8 & 67.9 & 55.1 & \textbf{83.5} & 74.1 & 59.3 & 84.8 & 72.5          \\
    PLUE\cite{wang2022source}         & 51.1 & 75.3 & 77.8 & 69.2 & 72.8 & 75.5 & 66.3 & 49.2 & 77.2 & 73.4 & 54.0 & 81.2 & 68.6          \\
\hline
    \textbf{EBPR}  & 57.0 & \textbf{79.3} & \textbf{84.1} & 68.6 & 78.3 & \textbf{79.6} & \textbf{69.0} & \underline{56.8} & \underline{83.4} &\underline{75.1} & \underline{61.1} & \textbf{85.5} & \textbf{73.2}\\	
\hline
\end{tabular}
}
\vspace{-0.4cm}
\end{table*}

\begin{table*}[ht]
\caption{\footnotesize Classification accuracy (\%) on the \textbf{VisDA-C} dataset.}\label{VisDA-C}
\resizebox{\textwidth}{!}{
\begin{tabular}{|l|l|l|l|l|l|l|l|l|l|l|l|l|l|}
\hline
    Methods  & plane & bike & bus & cat & horse & knife & mcycle & person & plant & sktbrd & train & truck & Avg. \\
\hline
Source-only  & 67.9 & 11.1 & 57.7 & 70.9 & 63.3 & 8.9  & 79.1 & 21.6 & 68.1 & 16.8 & 84.6 & 9.4  & 46.6 \\
    \hline
DANN\cite{ganin2016domain}        & 81.9 & 77.7 & 82.8 & 44.3 & 81.2 & 29.5 & 65.1 & 28.6 & 51.9 & 54.6 & 82.8 & 7.8  & 57.4 \\
    CDAN\cite{long2018conditional}       & 85.2 & 66.9 & 83.0 & 50.8 & 84.2 & 74.9 & 88.1 & 74.5 & 83.4 & 76.0 & 81.9 & 38.0 & 73.9 \\
    SWD\cite{lee2019sliced}         & 90.8 & 82.5 & 81.7 & 70.5 & 91.7 & 69.5 & 86.3 & 77.5 & 87.5 & 63.6 & 85.6 & 29.2 & 76.4 \\
    MCC\cite{jin2020minimum}        & 88.7 & 80.3 & 80.5 & \underline{71.5} & 90.1 & 93.2 & 85.0 & 71.6 & 89.4 & 73.8 & 85.0 & 36.9 & 78.8 \\
    PAL\cite{hu2020panda}         & 90.9 & 50.5 & 72.3 & \textbf{82.7} & 88.3 & 88.3 & \textbf{90.3} & 79.8 & 89.7 & 79.2 & \underline{88.1} & 39.4 & 78.3 \\
\hline
SHOT\cite{liang2020we}       & 94.3 & \textbf{88.5} & 80.1 & 57.3 & 93.1 & 94.9 & 80.7 & 80.3 & 91.5 & 89.1 & 86.3 & 58.2 & 82.9 \\
SFDA\cite{kim2021domain}       & 86.9 & 81.7 & \underline{84.6} & 63.9 & 93.1 & 91.4 & 86.6 & 71.9 & 84.5 & 58.2 & 74.5 & 42.7 & 76.7 \\
    BAIT\cite{yang2020unsupervised}        & 93.7 & 83.2 
    & 84.5 & 65.0 & 92.9 & \textbf{95.4} & 88.1 & 80.8 & 90.0 & 89.0 & 84.0 & 45.3 & 82.7 \\
    A2Net\cite{xia2021adaptive}        & 94.0  & \underline{87.8} & \textbf{85.6} & 66.8 & 93.7 & \underline{95.1} & 85.8 & 81.2 & 91.6 & 88.2 & 86.5 & 56.0 & \underline{84.3} \\
    HCL\cite{huang2021model}        & 93.3 & 85.4 & 80.7 &68.5 & 91.0 & 88.1 & 86.0 & 78.6 & 86.6 & 88.8 & 80.0 & \textbf{74.7} & 83.5 \\
    AAA\cite{li2021divergence}         & 94.4 & 85.9 & 74.9 & 60.2 & \underline{96.0} & 93.5 & 87.8 & 80.8 & 90.2 & \underline{92.0} & \underline{86.6} & \underline{68.3} & 84.2 \\
    DIPE\cite{wang2022exploring}       & \underline{95.2} & 87.6 & 78.8 & 55.9 & 93.9 & 95.0 & 84.1 & \underline{81.7 }& \underline{92.1} & 88.9 & 85.4 & 58.0 & 83.1 \\
    PLUE\cite{wang2022source}       & 89.4 & 82.7 & 83.5 & 61.2 & 92.8 & 92.1 & 86.0 & 79.8 & 69.0 & 69.0 & 79.6 & 44.1 & 78.9 \\
\hline
    \textbf{EBPR} & \textbf{95.9} & 86.5 & 84.4 & 65.2 
    & \textbf{96.1} & 93.1 & \underline{90.0} & \textbf{82.6} & \textbf{92.5} & \textbf{93.7} & \textbf{89.9} & 56.9 & \textbf{85.6}	\\			
\hline

\end{tabular}
}
\vspace{-0.2cm}
\end{table*}

\begin{table}[ht]
    \centering
        \caption{\footnotesize Ablation study on the modules.}
        \begin{tabular}{|l|l|l|l|l|}
      \hline
        Backbone & PGTC & AET & ECT & Avg.\\
        \hline
        \checkmark &   &  &   & 59.6      \\
        \checkmark & \checkmark &  &   & 70.6 \\
        \checkmark & \checkmark & \checkmark &  & 72.0  \\
        \checkmark & \checkmark & \checkmark & \checkmark & \textbf{73.2}  \\
        \hline
    \end{tabular}
    \label{tab5}
    
\end{table}

\begin{figure}[h]
    \centering
    \includegraphics[width=0.4\textwidth]{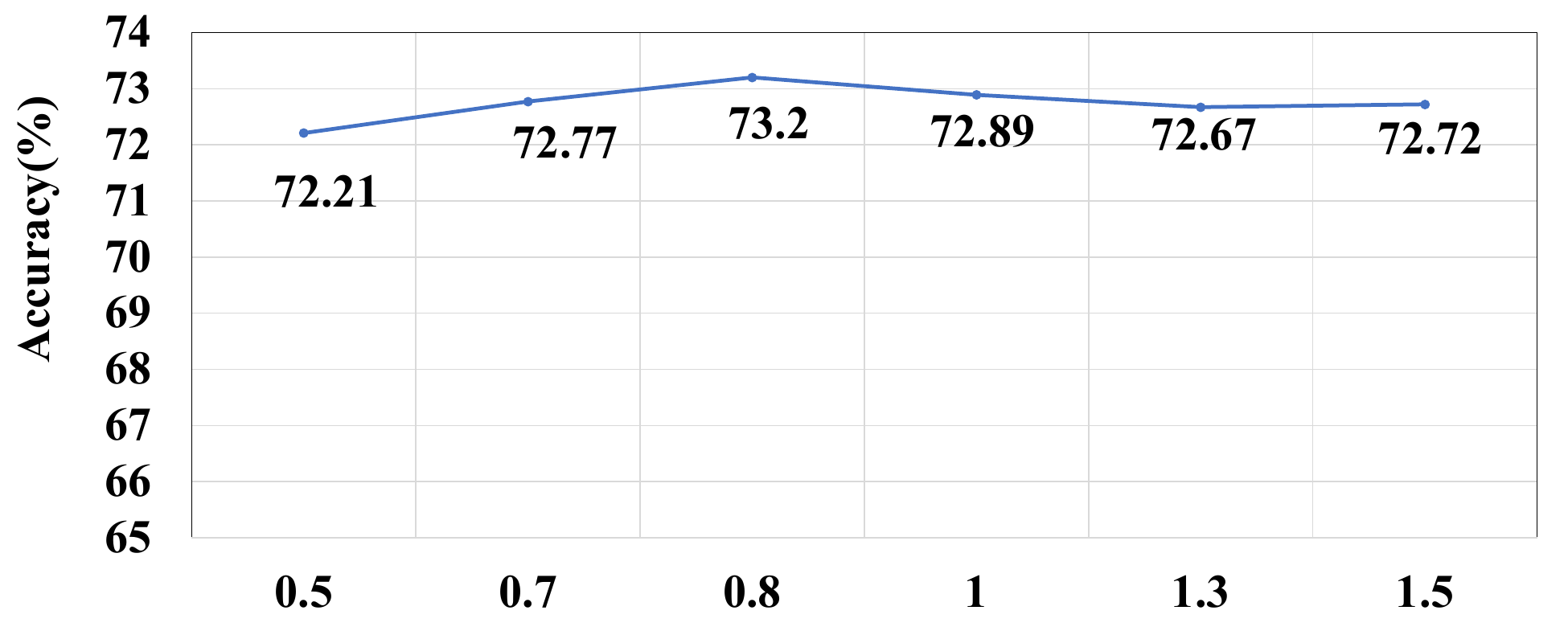}
\caption{\footnotesize Influence of  $\rho$ in C → A accuracy (\%) on the Office-Home dataset.} \label{fig2}
\vspace{-0.6cm}
\end{figure}

 Table \ref{office-31} shows the performance of EBPR on the Office-31 dataset. On the six tasks of the dataset, EBPR achieved one of the most advanced average accuracy and was 0.3\% better than the next best method SRDC. This is due to the low difficulty of the migration tasks for various domains in the Office-31 dataset, and EBPR generates reliable pseudo-labeled samples through energy scores. In relatively simple tasks, it is easy to obtain a large number of reliable samples for training, and effectively adapt to the target domain data. For challenging domain migration tasks, the combination of category energy threshold and global energy threshold in EBPR can greatly reduce the negative migration caused by the accumulation of false false label noise, and greatly improve the adaptability of the target model.

 As shown in table \ref{Office-Home}, EBPR similarly achieves the state-of-the-art classification average accuracy (73.2\%) on the Office-Home dataset. This is because in the process of pseudo-label filtering, EBPR adopts adaptive joint threshold to filter effectively solve the problem of category imbalance caused by clustering in the generation stage of pseudo-label, improve the classification accuracy of a small number of categories, and effectively improve the performance of the target model. In addition, EBPR takes into account the knowledge of the target data in an approach that enhances consistency, considering that the target data structure contained in the difficult sample is relatively difficult to mine.

 Table \ref{VisDA-C} shows the evaluation experiments of EBPR on the VisDA-C dataset. The VisDA-C dataset has a large number of images and a large domain difference between the synthetic data and the real data, which is a great challenge for SFDA. For the method based on pseudo-labels, whether the internal structure information of the target data can be mined and the quality of the pseudo-labels is the key. When only the source domain model is used to challenge the domain transfer task, the average class accuracy is only 48.0\%, and the recognition performance on some categories such as "knife" and "truck" is extremely low, resulting in a serious class performance imbalance problem. The EBPR method presented significantly mitigated this problem. In the challenging categories of "person" and "sktbrd", EBPR methods reached 82.6\% and 93.7\% respectively, 1.9\% and 4.6\% higher than SHOT. The inconsistency of category information caused by domain migration still exists to a certain extent, so it does not achieve the most advanced effect for "truck" class and "bicycle" class. However, EBPR effectively improves the generation and selection process of pseudo-labels, and fully considers the exploration of the structure information of the target domain. Our method obtains the optimal performance on the whole, which proves the effectiveness of this method on large data sets.



\begin{figure*}[t]
    \centering
	\begin{subfigure}{0.3\textwidth}
		\centering
		\includegraphics[width=0.75\linewidth]{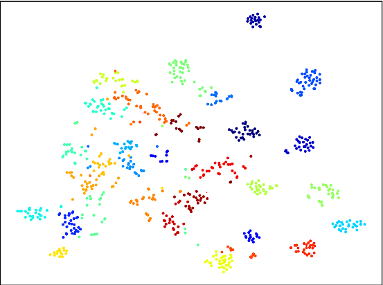}
		\caption{\footnotesize Source-only}
		\label{fig5 a}
	\end{subfigure}
	\centering
	\begin{subfigure}{0.3\linewidth}
		\centering
    		\includegraphics[width=0.78\linewidth]{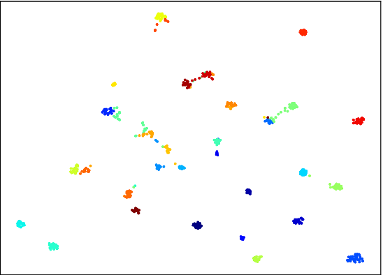}
		\caption{\footnotesize SHOT}
		\label{fig5 b}
	\end{subfigure}
	\centering
	\begin{subfigure}{0.3\linewidth}
		\centering
    		\includegraphics[width=0.73\linewidth]{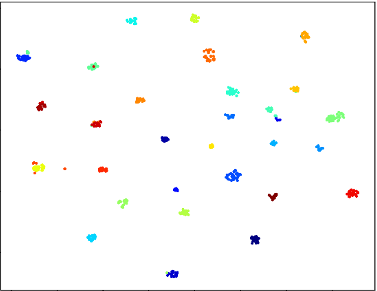}
		\caption{\footnotesize EBPR}
		\label{fig5 c}
	\end{subfigure}
	
	\vspace{2mm}
	\begin{subfigure}{0.3\linewidth}
		\centering
		\includegraphics[width=0.78\linewidth]{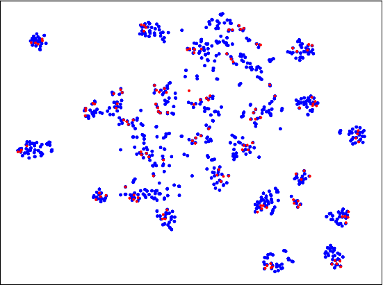}
		\caption{\footnotesize Source-only}
		\label{fig5 d}
	\end{subfigure}
	\centering
	\begin{subfigure}{0.3\linewidth}
		\centering
    		\includegraphics[width=0.80\linewidth]{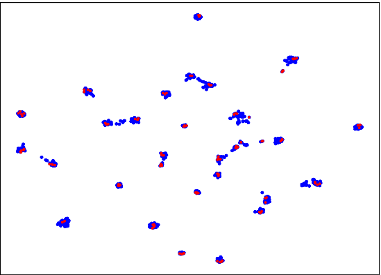}
		\caption{\footnotesize SHOT}
		\label{fig5 e}
	\end{subfigure}
	\centering
	\begin{subfigure}{0.3\linewidth}
		\centering
    		\includegraphics[width=0.75\linewidth]{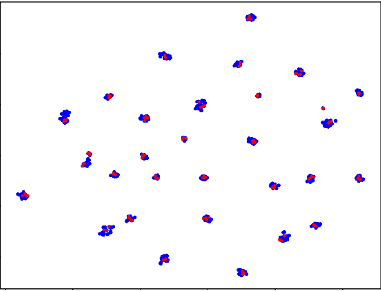}
		\caption{\footnotesize EBPR}
		\label{fig5 f}
	\end{subfigure}
	\caption{\footnotesize The t-SNE of the task A → W. The first row shows the visualization of category features where the different colors denote different categories. The second row shows features of the filtered intermediate samples (in red) and others (in blue).}
	\label{fig333}
 \vspace{-0.3cm}
\end{figure*}

\subsection{Critical analysis}
\subsubsection{Component analysis}
To investigate the effectiveness of the proposed modules in the adaptation phase, the quantitative results of the model with different components are shown in Table \ref{tab5}. Each module in EBPR contributes uniquely, and the best results are achieved when all three modules are combined, synergistically enhancing the adaptability of the target model.

\subsubsection{Sensitivity to hyper-parameters}
 Fig.\ref{fig2} illustrates the hyper-parameter $\rho$ sensitivity variation of EBPR on the Office-Home dataset. The model achieves the best classification effect. Finally, EBPR sets $\rho$=0.8.

\subsubsection{t-SNE visualization}
The target features exhibit clear clustering in Fig.\ref{fig333}. Comparing images (c) and (e), it is evident that EBPR reduces inter-category adhesion compared to SHOT, resulting in a more pronounced clustering effect. Similarly, comparing images (d) and (f), EBPR demonstrates better feature alignment between the source and target domains, highlighting its effectiveness in addressing deep domain adaptation challenges of SFDA. The energy score can be used to measure the difference between the source domain and the target domain. The smaller the difference between the target domain samples with lower energy and the source domain data samples, the model can better adapt to the target domain by minimizing the energy score of the target domain samples. Setting a threshold on the energy score can implicitly classify the target domain sample as "adaptive" or "unadaptive". The energy-based model can improve the accuracy by exploring the class similarity and sample similarity between samples in the energy layer. Our work demonstrates that energy-based models can attract more attention in more relevant fields in the future.

\section{Conclusion and limitation}
We proposed a method of SFDA based on energy and optimal transport named EBPR, which effectively addresses challenges related to target domain structure information neglect and noise pseudo-labels in the training process. By employing energy-based adaptive joint thresholds for pseudo-label filtering and integrating enhanced consistency training to capture the intrinsic structure of challenging samples, EBPR consistently demonstrates superior performance and effectiveness across diverse SFDA benchmark datasets.  EBPR has better performance, but its limitation may be that the time cost will be greater when the data set is too large, which is also a common limitation in the field of SFDA.

\printcredits












\end{document}